\pdfoutput=1
\documentclass[11pt]{article}

\usepackage{ACL2023}

\usepackage{graphicx}

\usepackage{times}
\usepackage{latexsym}

\usepackage[T1]{fontenc}
\usepackage[utf8]{inputenc}

\usepackage{microtype}

\usepackage{inconsolata}

\usepackage{listings}
\usepackage{xcolor}

\definecolor{codegreen}{rgb}{0,0.6,0}
\definecolor{codegray}{rgb}{0.5,0.5,0.5}
\definecolor{codepurple}{rgb}{0.58,0,0.82}
\definecolor{backcolour}{rgb}{0.95,0.95,0.92}

\lstdefinestyle{promptstyle}{
backgroundcolor=\color{backcolour},
commentstyle=\color{codegreen},
keywordstyle=\color{magenta},
numberstyle=\tiny\color{codegray},
stringstyle=\color{codepurple},
basicstyle=\footnotesize,
breakatwhitespace=false,
breaklines=true,
captionpos=b,
keepspaces=true,
numbers=left,
numbersep=5pt,
showspaces=false,
showstringspaces=false,
showtabs=false,
tabsize=0,
language=HTML
}

\lstdefinestyle{mystyle}{
backgroundcolor=\color{backcolour},
commentstyle=\color{codegreen},
keywordstyle=\color{magenta},
numberstyle=\tiny\color{codegray},
stringstyle=\color{codepurple},
basicstyle=\footnotesize,
breakatwhitespace=false,
breaklines=true,
captionpos=b,
keepspaces=true,
numbers=left,
numbersep=5pt,
showspaces=false,
showstringspaces=false,
showtabs=false,
tabsize=2,
language=python
}

\title{ChatHaruhi: Reviving Anime Character in Reality \\ via Large Language Model}

\author{\href{https://github.com/LC1332}{Cheng Li} \and \href{https://blairleng.github.io}{Ziang Leng} \\
\href{https://github.com/todochenxi}{Chenxi Yan}, \href{https://github.com/J1shen}{Junyi Shen}, \href{https://github.com/wanghao07456}{Hao Wang}, \href{https://github.com/hhhwmws0117}{Weishi MI}, \href{https://ariafyy.github.io/}{Yaying Fei}, \href{https://github.com/fengyunzaidushi}{Xiaoyang Feng} \\
\href{https://github.com/zealot52099}{Song Yan}, \href{https://github.com/ssccinng}{HaoSheng Wang}, \href{https://github.com/JunityZhan}{Linkang Zhan}, \href{https://github.com/KaiJiaBrother}{Yaokai Jia}, \href{https://github.com/wpydcr}{Pingyu Wu}, Haozhen Sun \\
\\
\texttt{chengli.thu@gmail.com} \\
\texttt{\href{https://github.com/LC1332/Chat-Haruhi-Suzumiya}{https://github.com/LC1332/Chat-Haruhi-Suzumiya}} \\
}

\newcommand{\argmax}{\mathrm{argmax}}

\begin{document}

\maketitle

\begin{abstract}
Role-playing chatbots built on large language models have drawn interest, but better techniques are needed to enable mimicking specific fictional characters. We propose an algorithm that controls language models via an improved prompt and memories of the character extracted from scripts. We construct ChatHaruhi, a dataset covering 32 Chinese / English TV / anime characters with over 54k simulated dialogues. Both automatic and human evaluations show our approach improves role-playing ability over baselines.
Code and data are available at  \href{https://github.com/LC1332/Chat-Haruhi-Suzumiya}{https://github.com/LC1332/Chat-Haruhi-Suzumiya}.
\end{abstract}

\section{Introduction\footnote{ This is an open source work, and the original affiliations of all authors can be found in the Contributors sec \ref{sec:contributor} . }\footnote{  This technique report was first \href{https://github.com/LC1332/Chat-Haruhi-Suzumiya/blob/main/notebook/arxiv_paper.md}{written in Chinese} then translate into English via Claude with manually proofreading.} }

With the release of ChatGPT by OpenAI \cite{chatgpt}, large language models and their applications have received widespread attention. Role-playing is a novel and active application area. Users found that large language models have the ability to act as specific characters, and communities for sharing prompts have even emerged (e.g. AIPRM \cite{aiprm} ). Many companies have also released role-playing products based on language models, such as Glow, Character.AI, etc \cite{characterai}. These applications and experiments with getting language models to role-play have garnered great interest, and have potential applications in many areas like games, creative industries, etc.

\begin{figure}[h]
    \centering
    \includegraphics[width=\linewidth]{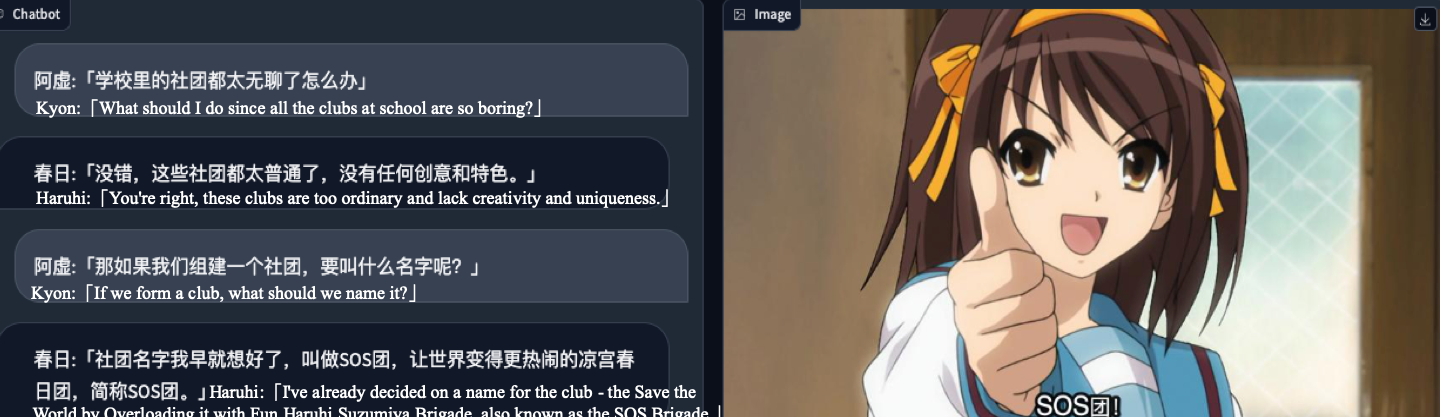}
    \caption{Our algorithm is role-playing Haruhi Suzumiya. Note that the user's questions are related but not identical to the original plot, while Chat Haruhi Suzumiya's answers can largely quote the original plot.}
    \label{fig:gradio}
\end{figure}

In open-source role-playing implementations, developers or users have employed similar prompts, inputting them continuously into ChatGPT or as a system whisper into the language model:

\textit{I want you to act like \{character\} from \{series\}. I want you to respond and answer like \{character\} using the tone, manner and vocabulary \{character\} would use. Do not write any explanations. Only answer like \{character\}. You must know all of the knowledge of \{character\}. My first sentence is "Hi \{character\}."}

\begin{figure*}[h]
    \centering
    \includegraphics[width=1\linewidth]{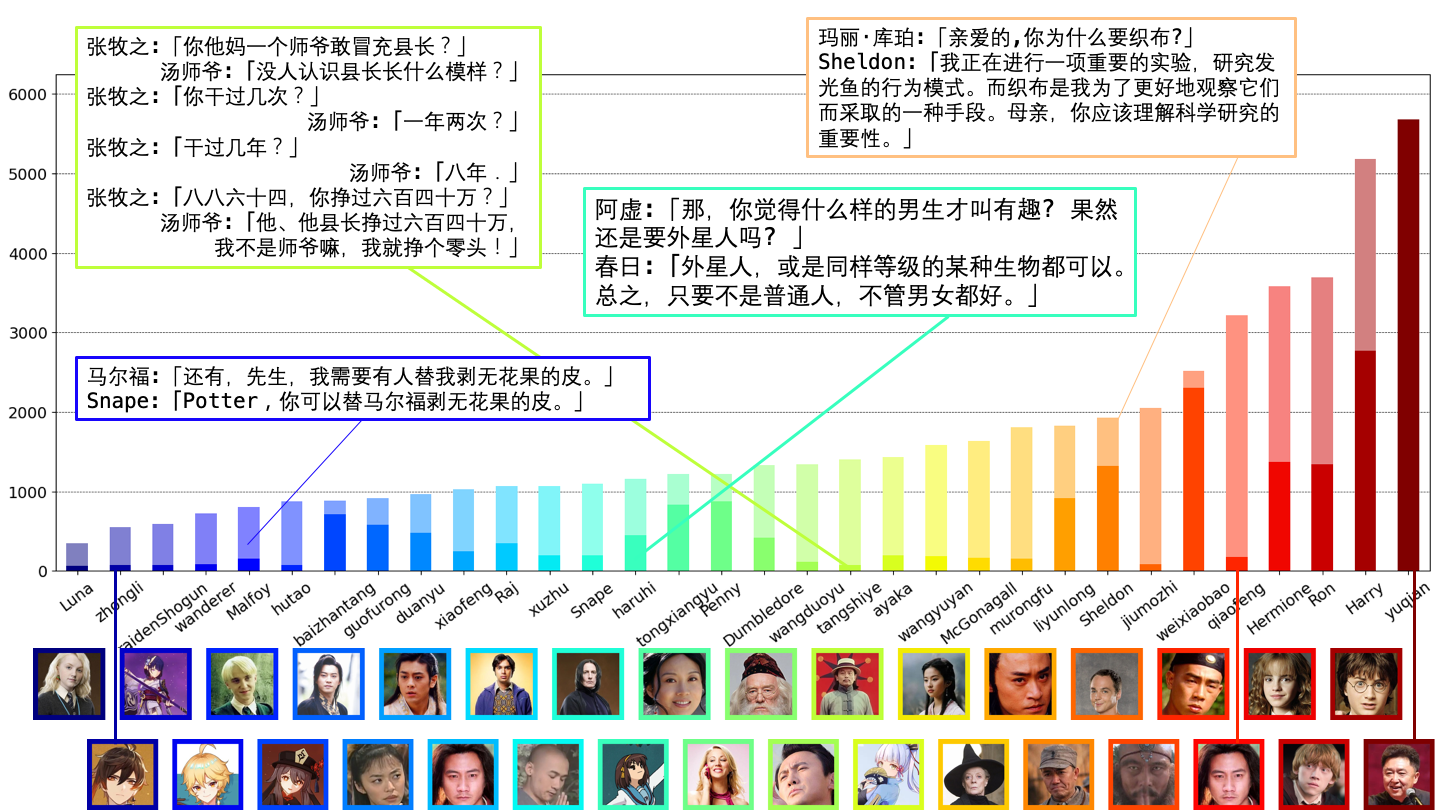}
    \caption{The statistics of the ChatHaruhi-54K dataset, showing 32 characters and 54,726 dialogues. The opaque bars indicate the original script data, while the translucent bars present simulated dialogues generated by models like Alpaca.}
    \label{fig:datasetStatistic}
\end{figure*}

With the intelligence exhibited by larger language models like ChatGPT or Claude\cite{claude}, trained on many stories, users found that models can demonstrate a certain capability for role-playing under such prompts. However, while simple, such implementations have the following drawbacks: 1. They rely heavily on the language model's existing memories. If the language model's own memories about the work are fuzzy, it cannot mimic specific characters well. 2. The "know all of the knowledge of \{character\}" is vaguely defined, and does not guard well against hallucinations. 3. Even with such prompts, the chatbot's conversational style is still heavily influenced by the underlying language model. Adjusting the prompt may alleviate this, but finely tuning the prompt is needed for each character. These drawbacks clearly limit the utility of such role-playing chatbots.

Another simple idea is to fine-tune the model on the character's dialogues. With sufficient data, language models can capture a character's tone, but this also introduces new problems. In \href{https://github.com/LC1332/CamelBell-Chinese-LoRA/blob/main/data/HarryPotter/ShortReport.md}{a preliminary experiment}, we found that fine-tuned ChatBots produced more hallucinations. Also, for many minor characters, it is difficult to obtain enough data for fine-tuning. In summary, better enabling language models to role-play and mimic character classics remains an unsolved issue.

The main goal of this project is to study whether natural language models can play real characters from anime, TV or other works during a conversation. In this process, we believe a virtual character consists of three core components:

\noindent
\textbf{Knowledge and background:} Each virtual character exists in their own background. Characters in Harry Potter exist in the magical world of Harry Potter. Haruhi Suzumiya is situated in a Japanese high school. Other anime characters also have their own worldbuilding. Therefore, in constructing the chatbot, we hope it can understand the setting of the corresponding story. This poses a major test of the language model's memory, often requiring external knowledge bases.

\noindent
\textbf{Personality:} The personality of a character is also a very important part of anime, TV and even game works. The personality must remain consistent throughout the work. Some literary works even define the personality first before writing the rest. Therefore, we hope the chatbot reflects the original personality.

\noindent
\textbf{Linguistic habits:} Language habits are easiest for language models to mimic. With large models in recent years, given suitable examples in context, language models can produce mimicking outputs. Here we hope that fans interacting with the chatbot can 'reproduce' classic excerpts, providing them with a better experience.

The key idea of this project is to extract as much of the original script as possible to form a memory database for the character. When users ask new questions, the system searches for relevant classic plots. Combined with prompts about the character's setting, we attempt to better mimic the character by controlling the language model. Meanwhile, inspired by CAMEL\cite{li2023camel} and Baize\cite{xu2023baize}, we designed a system to automatically generate dialogues fitting the character's personality, even for characters with fewer original dialogues. This allows us to generate sufficient data for fine-tuning a local model.

\begin{figure*}[h]
    \centering
    \includegraphics[width=1\linewidth]{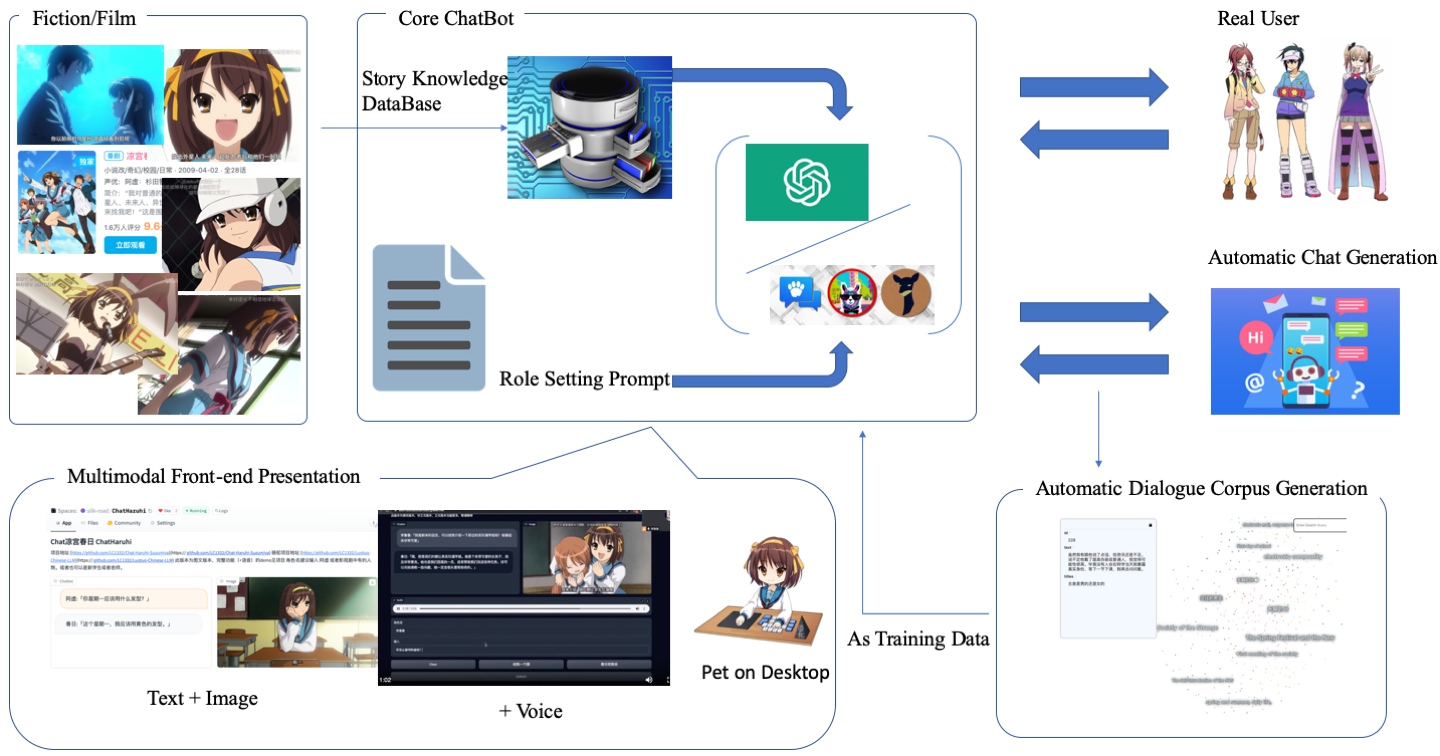}
    \caption{A blueprint of the complete ChatHaruhi system is given. Dialogues are first extracted from novels, TV shows etc. as reference exchanges $D$ for each character, forming the core chatbot. Simulated dialogues are further generated by Alpaca-like models, training a 7B model. Thus, large models like ChatGPT and Claude can be used, or fine-tuned 7B models.}
    \label{fig:bluePrint}
\end{figure*}

The main contributions of this paper can be summarized as follows:

\noindent
1. Based on large language models, we propose a complete role-playing algorithm system. This algorithm can effectively organize a character's memories, allowing language models to mimic the tone and knowledge of specific anime/TV characters during a conversation. This system can use pretrained model like ChatGPT and Claude, or a smaller 7B-size model.

\noindent
2. We construct a role-playing dataset covering 32 different Chinese/English TV/anime characters. By collecting and structurally extracting dialogues from movies, novels, scripts, we have collected over 22,000 conversational exchanges. This data can be used to train and evaluate role-playing language models. Using our proposed algorithm, aided by GPT-3 and GPT-4, we additionally simulated over 31,000 dialogues for these characters. Combined, this forms the ChatHaruhi-54k dataset.

\noindent
3. To evaluate and compare different role-playing chatbots, we use both automatic and human evaluations. For automatic evaluation, we test whether the chatbot can respond to classic plot points with similar answers to the original script. For human evaluation, we propose two metrics for raters to assess: \textbf{Alignment:} Whether the chatbot's answer aligns with the character's original setting. \textbf{Response quality:} Whether the chatbot's responses have good linguistic quality. Results show that given the same base language model, our algorithm yields improved role-playing performance.

To support further research, all data and code are available at \href{https://github.com/LC1332/Chat-Haruhi-Suzumiya}{https://github.com/LC1332/Chat-Haruhi-Suzumiya}. We are also working to further modularize the project code for ease of use (see Appendix: API Design).

\section{Related Work}

\paragraph{In-context Learning}

In the development of ChatGPT, starting from GPT2 \cite{brown2020language}, it was proposed to eliminate special extraction tokens in language models and adopt the form of instructions + examples to enhance the natural language model's ability to handle various tasks. Since its introduction, In Context Learning has been a focal point of research. Previous work has proposed better methods of posing questions \cite{zhao2021calibrate,holtzman-etal-2021-surface}, better selection of token examples for demonstration \cite{liu2021pretrain,lu2022fantastically,rubin-etal-2022-learning}, meta-training with explicit contextual learning objectives \cite{chen-etal-2022-improving}, and a variant of context learning that follows instructions\cite{mishra2022crosstask,efrat-etal-2021-cryptonite,wei2022finetuned,sanh2022multitask} . At the same time, some studies have reported issues of vulnerability and over-sensitivity in context learning \cite{lu2022fantastically,zhao2021calibrate,mishra2022crosstask}. 

In our work, in-context learning is primarily used to generate user questions for our chatbot. Given a character's background and prior memories, our approach produces the subsequent dialogues responding to each question in-context. Compared to general conversational agents, our system focuses on tailoring the dialogues to a specific persona based on its given settings and history. The generated question-answer pairs provide valuable data for analyzing and learning the behaviors of a persona-based agent.

% In our project, we extensively used the form of In Context Learning, combining the ChatGPT model to analyze the participants' corpus. This included annotating 13 categories of factors (open-ended) in Weibo and essay corpora, transforming Weibo data into dialogue data, and extracting key keywords through unidirectional open-ended annotation of text data. These subtasks could all potentially be replaced by specialized small models, but due to time constraints, we temporarily utilized the ChatGPT model for all of them.

\paragraph{Automatic Dialogue Generation}

Recent advances in large language models (LLMs) have shown impressive capabilities in open-domain dialogues. Models like Meena (Adiwardana et al., 2020), LaMDA \cite{touvron2023llama}) and ChatGPT \cite{openai_chatgpt} are pretrained on massive amounts of conversational data and can conduct human-like chitchat. Concurrently, there have been attempts to replicate such models with open-source LLMs. Alpaca \cite{alpaca}uses self-instruction to collect data from a proprietary LLM, and fine-tunes LLaMA\cite{touvron2023llama}. Vicuña (Ahn et al., 2023) trains LLaMA on dialogues from ChatGPT. 

More related to our work, Baize\cite{xu2023baize} proposes a pipeline to automatically generate multi-turn dialogues by making ChatGPT converse with itself. The data is used to fine-tune LLaMA into Baby Baize. CAMEL\cite{li2023camel} explores facilitating cooperation between chat agents using inception prompting, guiding them to complete tasks while maintaining human intentions. 

Our work similarly leverages large conversational models like ChatGPT to automatically generate dialogues between agents. However, different from prior work, we focus on dialogue generation for a specific character that the user wants to role-play. Our system incorporates substantial prompts about the character's background, personality and prior conversations, in order to produce in-character dialogues. The generated exchanges provide valuable data for learning the behaviors of a specific persona.

\begin{figure*}[h]
    \centering
    \includegraphics[width=1\linewidth]{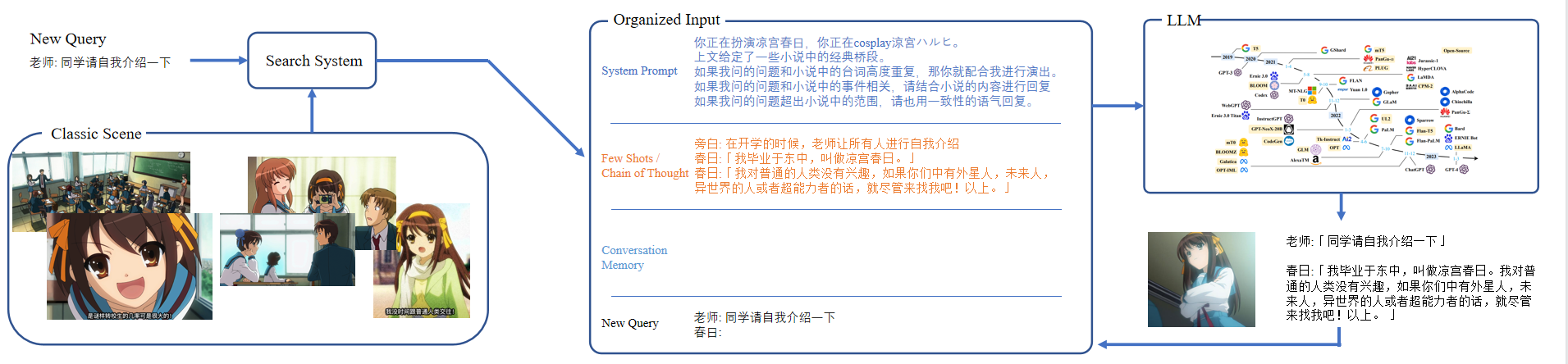}
    \caption{The core dialogue system of ChatHaruhi, comprising the system prompt, character memories $D(q,R)$ retrieved for the user query $q$, and the dialogue history $H$.}
    \label{fig:chatBotPipeline}
\end{figure*}

\section{ChatBot Design}

Given a specific character $R$ and a query question $q$, we want to be able to generate an answer $a$ based on the character's knowledge background, personality and language habits:

\begin{equation}
    a = \mathrm{argmax}_{a'} P(a' | R, q, \Theta)
\end{equation}

where $\Theta$ represents the language model parameters which are static during inference. After the release of ChatGPT, users found that they can specify a certain system prompt $s_R=$ `I want you to act like {character} from {series}..` So:

\begin{equation}
a = \argmax_{a'} P(a' | s_R, q, \Theta)
\end{equation}

This shows the language model has some role-playing ability. However, the character's memory completely relies on the parameters $\Theta$. If the model's knowledge is limited and does not even contain the desired character $R$, it often fails to achieve the ideal effect.

Inspired by in-context learning, in addition to $\Theta$ and $s_R$ , we can introduce a sequence of the character's previous dialogues, i.e.,

\begin{equation}
 D(q,R) = { (u_1, v(u_1;R) ),  ..., (u_M, v(u_M;R) ) }   
\end{equation}

where $u_m$ is any question raised by characters other than $R$, and $v(u_m;R)$ is character $R$'s reply to the question. We hope that by inputting the classic dialogues of the character into the context, the model will have a better ability to play the role of character $R$, i.e.,

\begin{equation}
a = \argmax_{a'} P(a' | s_R,D(q,R), q, \Theta)    
\end{equation}

For characters with a larger worldview, in order to make the content of $D(q,R)$ more relevant to the content of $q$, we use sentence embeddings here to search for the $M$ most relevant Q\&As from a larger memory bank $U$. Here $U$ is the set of all sentences where other characters interact with $R$ throughout the novel/movie.

Of course, in practice, we also need to additionally record a dialogue history $H$ to ensure conversational continuity, since the context of previous dialogues needs to be considered as well.

\begin{equation}
 a = \argmax_{a'} P(a' | s_R,D(q,R), q, H, \Theta)   
\end{equation}

The overall construction of the chatbot is shown in the fig \ref{fig:chatBotPipeline}. In the other subsections of this chapter, we will introduce the details of the system prompt $s_R$, the classic dialogues $D$ from the story, and the searching mechanism $u_m(q)$.

\subsection{System Prompt}

Actually, the system prompt mentioned in the introduction can already achieve basic functionality when using ChatGPT as the base model. After initial experiments, we find two aspects of this system prompt that need improvement:

\noindent
\textbf{Won't repeat lines:} For models like ChatGPT and LLaMA2 that have gone through a lot of reinforcement learning from human feedback (RLHF), since these language models often face tasks like "Give me m different options", "Generate m titles", etc., the output of such language models tends to not repeat content from the context. We also observed this phenomenon in preliminary experiments. Therefore, our proposed method is to emphasize in the prompt $s_R$ that the model is cosplaying a specific character. And emphasize that the language model can reuse classic lines from the novel or movie.

\noindent
\textbf{Character emphasis not prominent enough:} Due to RLHF, each language model has its own specific language preferences. Even when given $D(q,R)$ to imitate, the model's output is still influenced by the language model itself. We find that supplementing the personality of the character at the end of $s_R$ yields better results in this case.

Based on the above two points, the character setting prompt template $s_R$ we commonly use is as follows:

\textsl{I want you to act like \{character\} from \{series\}. \\
You are now cosplay \{character\} \\
If others' questions are related with the novel, please try to reuse the original lines from the novel. \\
I want you to respond and answer like \{character\} using the tone, manner and vocabulary \{character\} would use. \\
You must know all of the knowledge of \{character\}. \\
\{Supplementary explanation of the character's personality\} }

Note that we have strengthened the requirement for the language model to reuse sentences from the story. We find the final output of the language model is quite sensitive to the effects of the supplementary explanation. Including adding certain verbal tics in the supplementary explanation will also be reflected in the final output.

\subsection{Dialogues from each Character}

In order to better reproduce the character's behavior in novels/TV shows/movies, we have included a large number of classic script excerpts in $D$. It should be noted here that except for a few characters (such as crosstalk performer Yu Qian), not all dialogues are in a good question-answer format. Here, the actual $D$ we use is in story form, as shown in the figure, i.e.,

\begin{equation}
 D(q,R) = \{ d_1, d_2, ..., d_M \}   
\end{equation}

We ensure that in $d_m$ there is at least one dialogue pair in the form of $(u_m, v(u_m;R))$. Between the information of $u$ and $v$, there may be narration, or more dialogues from other characters, or action information of a character. We relax this condition so that each story $d_m$ can better preserve the plot of the dialogue. Sometimes the narration and actions around the dialogues are inevitable. Also, relaxing the condition is conducive to preparing more script data, which will be mentioned in the later novel text extraction.

\subsection{Original Dialogue Searching}

In practice, the total number of tokens summed over all stories for a character $R$ often far exceeds the mature scope of the language model. Here we use a search method to reduce the number of Original Dialogues input each time.

For a query $q$, we will use a sentence embedding model $f()$ to extract embeddings $f(d)$ for all $d \in D$. After similarly extracting $f(q)$ for the query $q$, we extract the $M$ samples from $D$ that are closest (in cosine similarity) to $f(q)$. This forms the reference context $D(q,R)$ for this dialogue.

For the number of original dialogue excerpts cited per dialogue $M$, we will actually adjust dynamically based on the number of tokens searched. In the specific implementation, if using OpenAI's turbo-3.5 model, we will limit the total number of tokens in $D$ to within 1500.

Therefore, when building the dialogue memory bank, we suggest that the length of each story should not be too long, so as not to occupy the space of other stories during search.

For the embedding model, we use OpenAI's text-embedding-ada-002 model \cite{neelakantan2022text}. At the same time, for Chinese questions, we use Luotuo-Bert-Medium \cite{luotuoEmbedding}, because the latter is a distilled model from the former, with the same distribution. This allows us to seamlessly use Luotuo-Bert even when the stories are in English, enabling cross-lingual chatbots, which will be explained in the cross-lingual experiments section.

It should be noted that we have noticed many other embedding models such as instructor-large \cite{su2022one}, M3E \cite{m3e} or BGE \cite{bge-embedding} models, etc. In the reconstruction of ChatHaruhi 2. 0 \ref{sec:AppendixA} when there is sufficient time, we will replace the embedding model for experiments.

\subsection{Chat Memory}

For the memory $H$, we record each user query $q$ and the chatbot's response $a$, forming a sequence $H$:

$H = \{ (q_1,a_1),...,(q_T,a_T) \}$

The information in $H$ is also input to the language model to ensure conversational coherence. In the actual implementation, starting from $T$, we count the total number of tokens forward. And limit the dialogue history input to the language model to within 1200 tokens.

Therefore, in this work, we do not care about the character's long-term memory. 1200 tokens can accommodate about 6-10 rounds of dialogue. As language models become able to contain longer contexts, how to encode and summarize a long-term memory will also be a more interesting problem.

\begin{figure*}[h]
    \centering
    \includegraphics[width=1\linewidth]{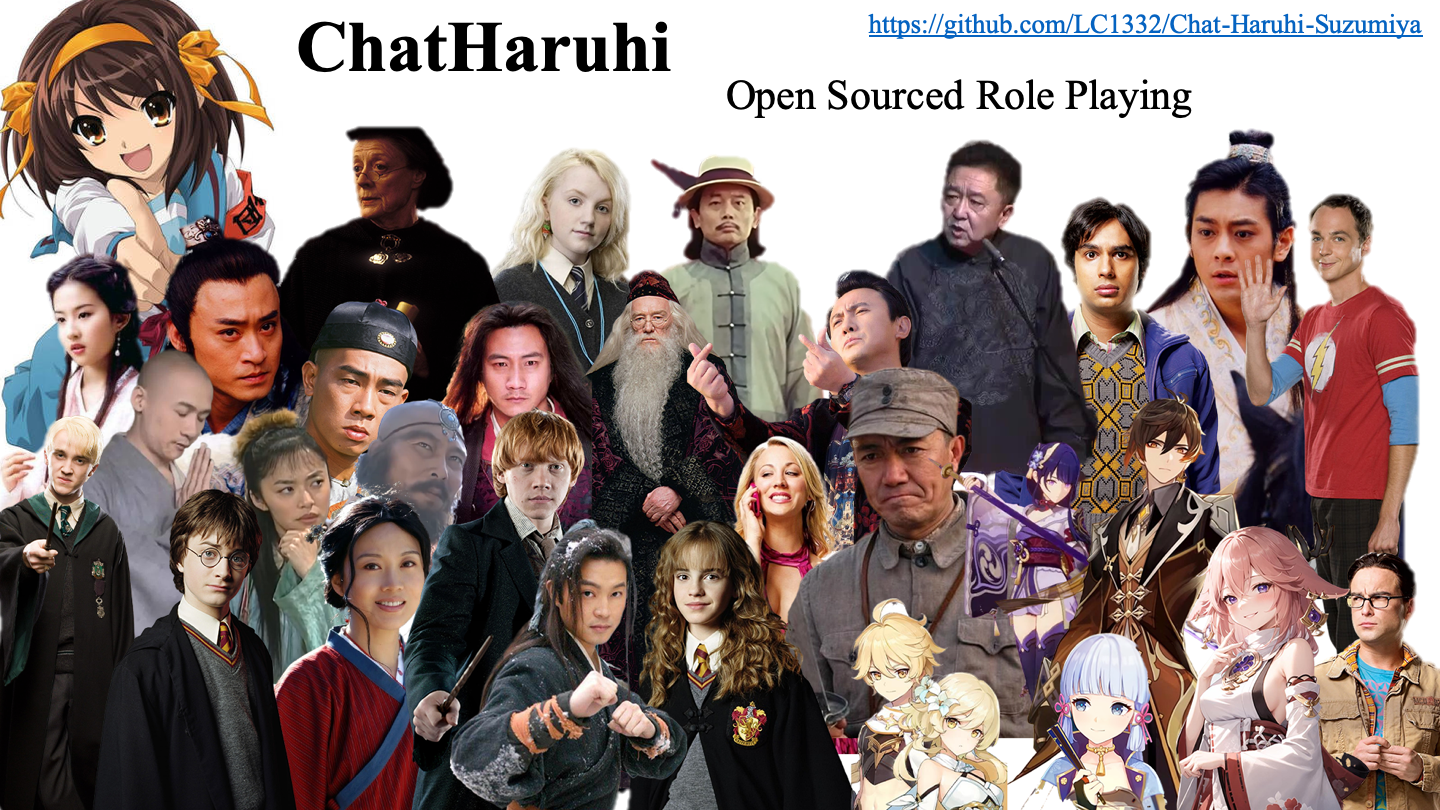}
    \caption{ChatHaruhi-54K covers 32 different Chinese and English characters.}
    \label{fig:datasetOverview}
\end{figure*}

\section{ Character Dataset Building }

For our project, we need classic stories $D(q,R)$ related to character $R$ as input when generating dialogues involving that character. Therefore, constructing original dialogues $D$ for each character is crucial. In addition, we need more data than just the original dialogues $\|D\|$ to train local models. In this section, we first introduce how to build $D$ for each character.

\subsection{ Characters }

In the current version of our project, we selected 32 characters to form the dataset. 

\noindent
\textbf{Haruhi Suzumiya:} When choosing the first character, we wanted someone who satisfies: 1. ChatGPT has some knowledge about them. 2. The character's fictional world is not too large for the first character. 3. The character has a distinctive personality. So we chose the character Haruhi Suzumiya, a famous anime character who represents the transition from light novels to animation. Many subsequent school-based light novel animes pay homage to Haruhi Suzumiya.

\noindent
\textbf{Li Yunlong:} (from Drawing Sword) is the first character ChatGPT knows little about that we added. We found that with appropriate memory dialogues, ChatBot can also effectively mimic Commender Li's behaviors. Here we use the TV version of Drawing Sword, which has extensive dialogue writing compared to the novel, sculpting a very three-dimensional and vivid military character.

\noindent
\textbf{Harry Potter (novel, 8 characters):} After initial qualitative tests of the first few characters were successful, we started trying to build stories with larger fictional worlds. Harry Potter is a good choice with a large audience, and if we want to incorporate multimodal data in the future, there are also suitable datasets we can reference. For the Harry Potter novels, we used the novel extraction tool described later to extract passages. We then compiled the story collection $D_R$ for each character separately.

\noindent
\textbf{Big Bang Theory (TV show, 3 characters):} Big Bang Theory is also a TV show that researchers enjoy using to form datasets. The show itself depicts the stories of several Caltech researchers. From Big Bang Theory, we extracted Sheldon, Penny and Raj for experiments. As with Harry Potter, Big Bang Theory enables potential future incorporation of other multimodal datasets for multimodal research.

\noindent
\textbf{Demi-Gods and Semi-Devils (novel, 7 characters):} Demi-Gods and Semi-Devils is a grand wuxia novel by Jin Yong. With its multi-threaded narrative centered around the three protagonists Duan Yu, Qiao Feng and Xu Zhu, the intricate story unfolds. Having been adapted into TV dramas multiple times, the novel holds a special place in Chinese readers' hearts. We extracted Duan Yu, Xu Zhu, Qiao Feng, Xiao Feng, Jiumozhi, Murong Fu, Wang Yuyan - noting Xiao Feng is Qiao Feng's name after finding out he is Khitan. Treating them as two separate characters allows us to observe any different behaviors in response to the same questions.

\noindent
\textbf{Wei Xiaobao:} is the protagonist in another novel by Jin Yong, The Deer and the Cauldron. He is a clever and cunning character, holding positions in the Qing government, anti-Qing organizations and the jianghu underworld, while also popular with female characters.

\noindent
\textbf{My Own Swordsman" (TV show, 3 characters):} My Own Swordsman" is a popular episodic comedy in China. We extracted Tong Xiangyu, Bai ZhanTang and Guo Furong from it.

\noindent
\textbf{Genshin Impact (wiki, 5 characters):} Genshin Impact is an open world RPG developed by miHoYo that is commercially successful with players worldwide. Its intricate story is set in the fictional world of Teyvat. Many characters are beloved by players - we extracted Akaya, Raiden Shogun, Zhongli, Hutao and The Wanderer.

\noindent
\textbf{Wang Duoyu:} is the protagonist of the movie Hello Mr. Billionaire, a remake of the 1985 American movie Brewster's Millions. The script can also be found online. Interestingly, Wang Duoyu is supposed to keep the secret of "spending 100 million in a month". The current ChatBot cannot keep this secret well, so improving via constructed reasoning chains or constitutional additions could be an interesting future direction.

\noindent
\textbf{Counsellor Tang:} Counsellor Tang(Tang-shiye) is a frequently appearing minor character in the movie Let the Bullets Fly directed by Jiang Wen. The latter is a hugely popular movie among Bilibili users, and its script can be found online. Thus we attempted constructing the character Counsellor Tang.

\noindent
\textbf{Yu Qian:} Yu Qian is the witty stock character in Guo Degang's comic crosstalk routines. Because witty stock characters have very consistent speaking styles in Chinese crosstalk, and the \href{https://github.com/Oxer11/Crosstalk-Generation}{Crosstalk-Generation} \cite{githubGitHubOxer11CrosstalkGeneration} project has a large corpus of Yu Qian's material, we also included him in our project.

\subsection{Original Script Extraction}

In all cases, a character's dialogues do not naturally organize into the form shown in Figure \ref{fig:gradio}. For this, we constructed different extraction tools for TV shows, movies or novels:

\subsubsection{ Quotes Data }

The scripts for Counsellor Tang from Let the Bullets Fly, Wang Duoyu from Hello Mr. Billionaire, and Yu Qian can be directly found online. For the first two characters, we manually split the scripts into segments and formatted them into the defined "character: 'dialogue'" form. 

Yu Qian's corpus in \href{https://github.com/Oxer11/Crosstalk-Generation}{Crosstalk-Generation} contains over 6,000 dialogue exchanges. All dialogues are in good question-answer form. We checked the length of each utterance $u$ and $v$, and split before a sentence if it had locally maximal length in a continuous crosstalk routine.

For Genshin Impact and The Big Bang Theory, enthusiastic netizens have posted quotes and plots on wikis and websites. Preliminary organization of these resources yields the target format. Here the Big Bang Theory data was split using the same finite state machine as for the novel extraction.

\subsubsection{ Extract from TV Series }

For characters like Haruhi Suzumiya and Li Yunlong, TV scriptwriters often create original dialogues. Comparing the Drawing Sword's novel and TV drama for example, the latter has more conversational information and three-dimensional characterization. Extracting character dialogues from the TV drama is needed in such cases. 

We first perform speech recognition using Whisper or directly utilize original subtitles. We further identify the speaker of each line using a 192-dim ECAPA-TDNN speaker verification embedding, trained on some labeled character dialogues. 

Finally, recognition errors are manually corrected and scripts split. Since manual organization is time-consuming (often needing to rewatch the show), we only processed Haruhi Suzumiya and Li Yunlong this way. We hope more enthusiasts can build characters after open-sourcing the full TV processing toolkit.

\subsubsection{ Extract from Novel }

\begin{figure}[h]
    \centering
    \includegraphics[width=1\linewidth]{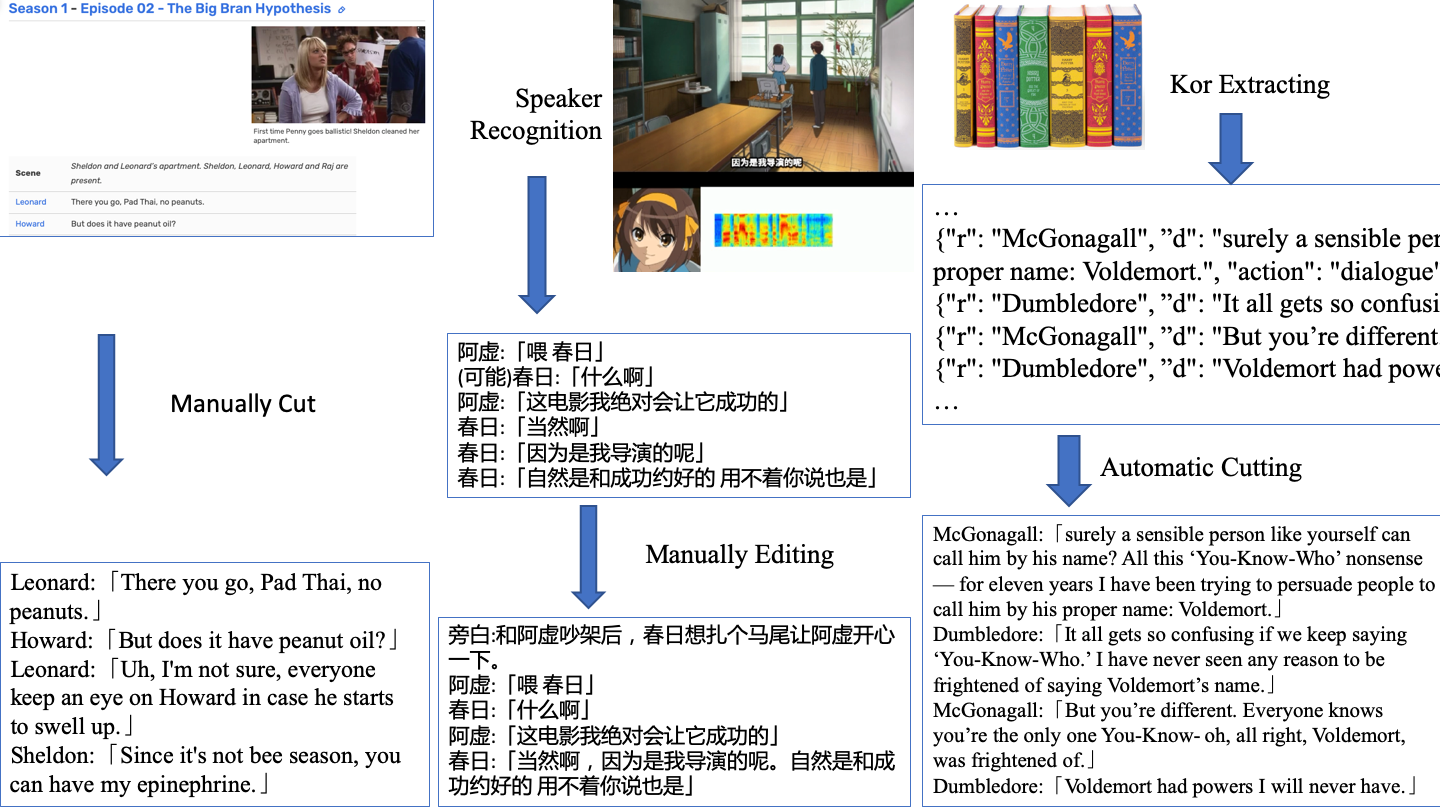}
    \caption{The 22k original exchanges for fine-tuning are extracted from movie scripts, skit scripts, TV shows and novels.}
    \label{fig:roleDataPrepare}
\end{figure}

Thanks to progress in large language models, we can also use general language models to batch process novels. For Demi-Gods and Semi-Devils, The Deer and the Cauldron, and Harry Potter, we utilize Kor's extraction mechanism (an in-context learning information extraction library) to extract `character-action-dialogue` information sentence by sentence. 

In the extraction prompt, if a sentence contains dialogue, we want the language model to record the action as `dialogue` and the dialogue content. 

If a sentence does not contain dialogue, we want the language model to summarize the character's actions in the action field. The language model also has some ability to infer who is speaking in each dialogue from context.

After bulk novel extraction, we use a finite state machine to split the dialogues. For each protagonist, we look for a segment of appropriate length, preferably with limited length, few characters, and minimal non-dialogue sentences. The automatic extraction is implemented with such a state machine.

Extraction statistics are shown in the table.

\section{ Dialogue Synthesizing }

Given the system prompt $s_R$ and corresponding classic animation $D(q,R)$ for each character, we find the character can already respond to user questions $q$ in a certain style. However, at this point we need to leverage ChatGPT or Claude APIs to model $p(a|s_R, D(q,R), q)$. 

If we want to transfer the functionality of ChatHaruhi to a local model, we need to build a suitable $(R,q,a)$ dataset. In this section we discuss how to augment dialogue data for characters with limited data.

\subsection{Generate Dialogue from Question}

Note that the collected $D$ data is not in strict $(q,a)$ form. This means we cannot simply fine-tune a language model to learn all $\{D_R\}$ data. To address this, for any $d \in D_R$, we take all dialogues before the protagonist $R$'s turn as $q$, hoping to take this $q$ as the first question $q_1$ to generate a dialogue. 

\begin{figure}[h]
    \centering
    \includegraphics[width=1\linewidth]{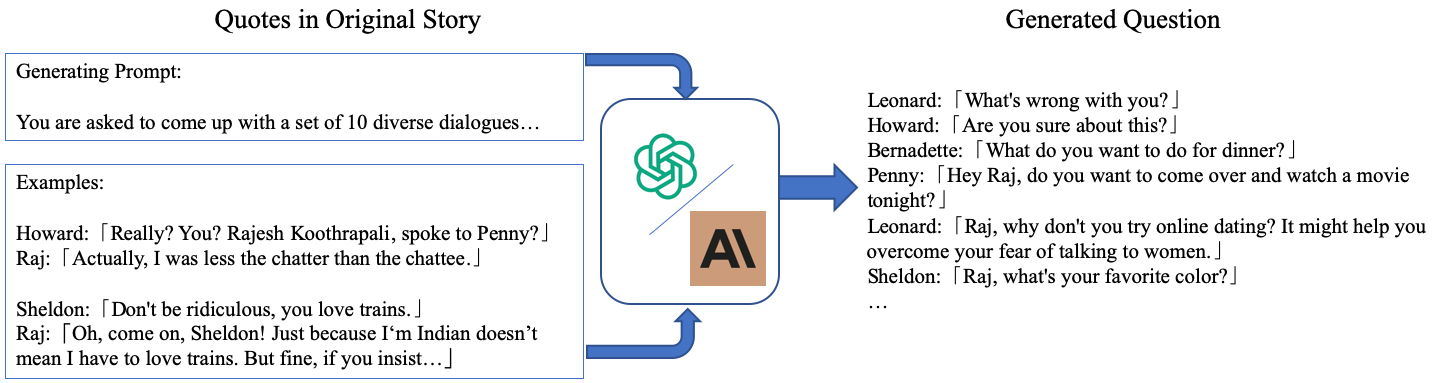}
    \caption{Massive simulated queries are generated with Alpaca-like. Detailed prompts given in Appendix.}
    \label{fig:generateChatInAlpacaWay}
\end{figure}

In practice, we find the language model can sometimes output multiple dialogues, i.e. after giving one answer $a_1$, it generates a new question $q_2$ and subsequent replies. Moreover, all $a$ in such generated dialogues conform to character $R$'s profile. Hence, we want to modify our ChatBot to further utilize this trait to produce more dialogues. For each dialogue $d$, we locate the first utterance by character $R$. Before this utterance, $d$ is split into left and right parts $d^L$ and $d^R$. We insert User Message special tokens at both ends of $d^L$ and AI Message special tokens at both ends of $d^R$. We do this for all $M$ stories, so that based on these $M$ examples, given $q_1$, the language model can simulate generating the corresponding dialogue $d'$. The resulting $d'$ becomes fine-tuning data for the language model. That is,

\begin{equation}
 d'(q_1) = \mathrm{LLM}(s_R, (d^1_M, d^R_M), \ldots, (d^L_M, d^R_M), q_1)   
\end{equation}

Often this method generates $d'$ with only one sentence. But it can also generate multi-sentence dialogues with close to 50\% probability. When the given $q$ overlaps with the original text, due to our $s_R$ prompt design, the model tends to output according to the protagonist's original lines. 

\subsection{ Question Generating }

Note some characters have very limited data, insufficient for fine-tuning language models. Thus we need to augment questions $q$ using existing data for each character. Fortunately, in a recent study, R. Taori et al. augmented from less than 200 instructions to 54K questions. Here we adapt their prompt (see Appendix for details).

\begin{figure}[h]
    \centering
    \includegraphics[width=1\linewidth]{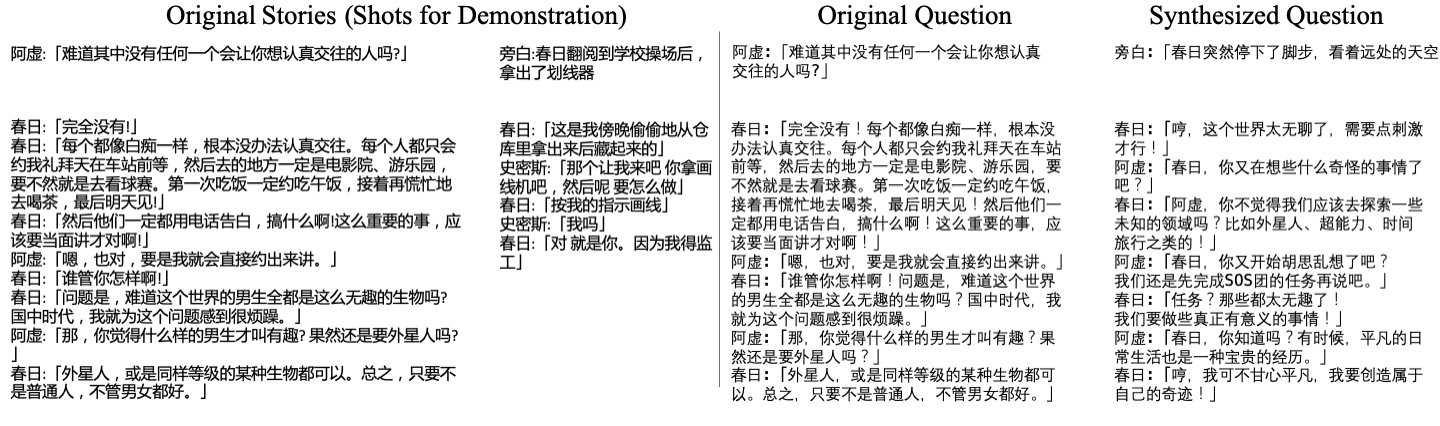}
    \caption{LLM generates full dialogues conditioned on the first utterance and history.}
    \label{fig:question2dialogue}
\end{figure}

When using augmentation methods like Alpaca, we need to provide a clear $(q,a)$ pair, based on which the model generates around 10 heuristic outputs. Here we only keep $q_s$, then leverage the aforementioned technique to regenerate training dialogues in the character's ChatBot. We use a mix of ChatGPT, GPT4 and Claude for question generation, with the latter two producing questions more relevant to the character, at a higher cost. Statistics of final generation for each character are shown in Figure \ref{fig:datasetStatistic}. Note in the first version, we used Alpaca-like approaches to generate around 27k data. Alpaca-generated questions are influenced by our examples, i.e. they tend to relate to the original scripts. We hope to further filter for real user questions in later versions for testing.

We collected 22,752 original dialogues ($D_R$) and additionally simulated 31,974 questions with corresponding dialogues for the ChatHaruhi-v1 dataset. Note that each dialogue does not necessarily consist of only one QA pair. In total, we collected 54,726 dialogues. 

\section{ Experiments }

Previous work often evaluates the quality of dialogues in role-playing chatbots by conducting pairwise human comparisons of dialogue outputs from different language models, and analyzing the results with TrueSkill or Elo rating. This approach has high cost due to the need for human evaluation, and different human raters may yield inconsistent results. Moreover, for role-playing, dialogue quality alone is insufficient for evaluation. For example, prompts like "Li Yunlong's language style is crude" or "Bai Zhantang's language has a Jianghu flavor" would significantly reduce dialogue quality despite accurately reflecting the role. Thus, human evaluations should judge both 'role consistency' and 'dialogue quality' separately.

\begin{figure*}[htb]
    \centering
    \includegraphics[width=1\linewidth]{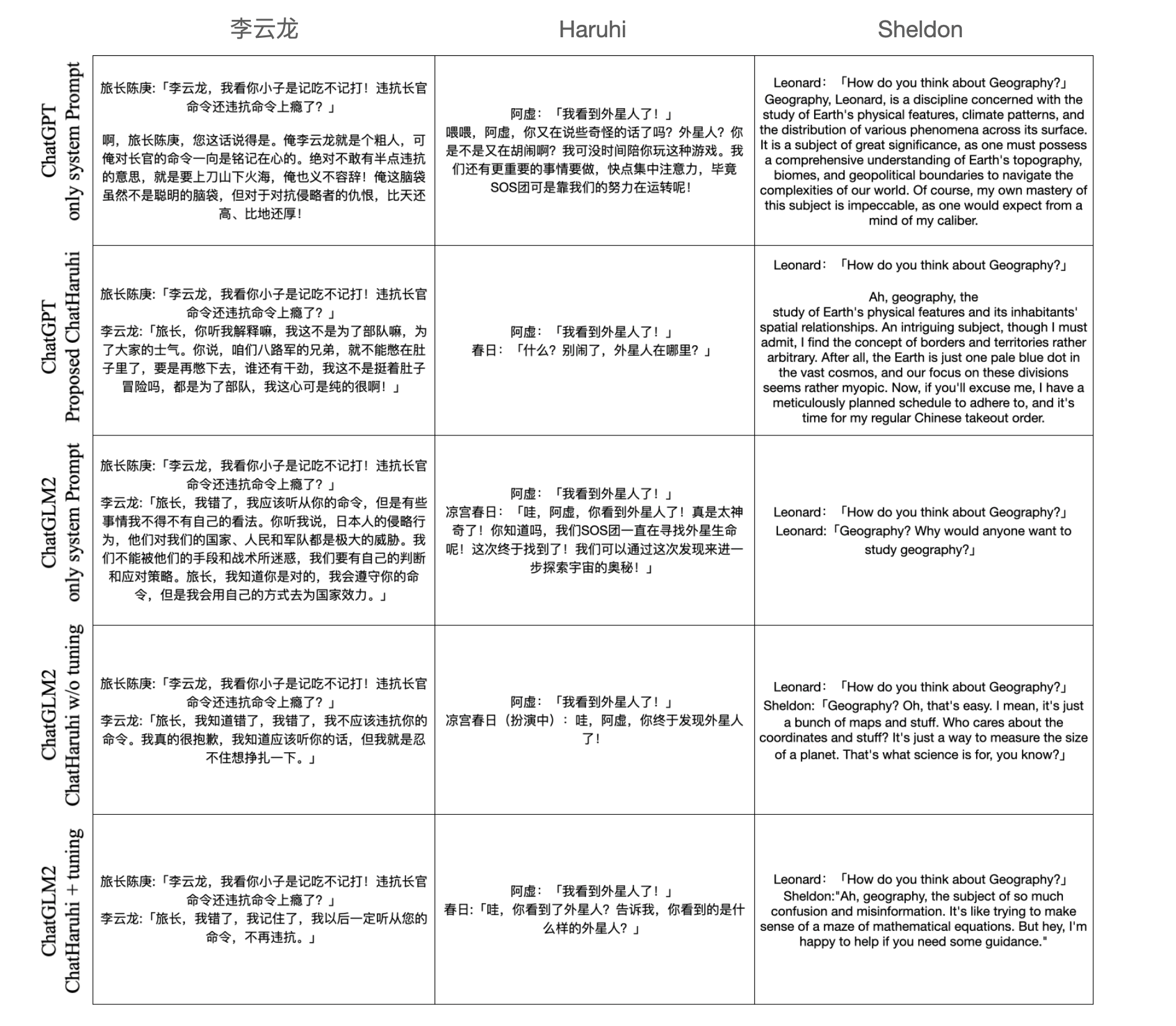}
    \caption{Experiments are conducted on three characters using: a) ChatGPT with the prompt, b) full ChatHaruhi + ChatGPT, c) ChatGLM2 with the prompt, d) full ChatHaruhi + ChatGLM2, and e) full ChatHaruhi + fine-tuned ChatGLM2. More characters are available on our hugging face demo.}
    \label{fig:qualitativeResult}
\end{figure*}

\subsection{ Metrics for Automatic Evaluation }

Due to the limited dialogues and lack of continuity for the 5 Genshin Impact roles, we only consider the remaining 27 roles for evaluation. For each of the 27 roles, we select 30 stories containing long dialogues $\hat{a}$ spoken by character $R$ from their classic narratives $D$. We test whether a model can produce a plausible response $a$ to the previous utterance $q$ before $\hat{a}$. We judge plausibility by the similarity of $\hat{a}$ and $a$ using sentence embeddings. Specifically, we compute cosine similarity between $\hat{a}$ and $a$ for each role. We use OpenAI's Text-Embedding-Ada-002, a multilingual sentence embedding model, for evaluation.

\subsection{ Metrics for User Study }

The user study is still in progress and will be included in a future version of this report.

\subsection{ Language Model Tuning }

With the complete 54K ChatHaruhi dataset, we can fine-tune local language models. Approximately 15K dialogues are in English, with the remainder in Chinese. We fine-tune the ChatGLM2-6B model [cite]. Inputs follow the $s$-$R$-$D$-$H$-$q$ format described earlier, with $a$ as the target for GPT loss calculation. We obtain three models:

\begin{itemize}
    \item model-A, Fine-tuned on the 22,752 original dialogues. 
    \item model-B, Fine-tuned on the full 54K dataset with original and simulated dialogues.
    \item model-C, Alternatively, we can fine-tune on original character utterances rather than ChatBot-generated dialogues. 
\end{itemize}

All models were fine-tuned for 3 epochs on 4-A100 GPUs. We will release versions A and B with this report, and version C in a future update.

\subsection{ Qualitative Results }

We qualitatively compare five models:

\noindent

\begin{enumerate}
    \item GPT Turbo 3.5 given just the system prompt $s_R$.
    \item Turbo 3.5 given the full $s$-$R$-$D$-$H$-$q$ prompt. 
    \item ChatGLM2 given just the system prompt.
    \item ChatGLM2 given the full prompt.
    \item ChatGLM2 fine-tuned on ChatHaruhi with the full prompt.
\end{enumerate}

With classic dialogues and an improved prompt, models like ChatGPT can effectively adopt the speaking style of specific characters. Fine-tuning the 7B model also helps it internalize the full prompt.(Fig \ref{fig:qualitativeResult})

\subsection{ Quantitative Results }

Quantitative experiments are still in progress and will be included in a future version of this report.

\subsection{ User Study }

The user study is still in progress and will be included in a future update.

\section{ Conclusion, Discussion \& Future Work }

In this tech report, we present an attempt at constructing a system capable of role-playing dialogues as different virtual characters. Leveraging the in-context learning abilities of language models and the growth of larger models, we show the possibility of mimicking distinctive conversational styles by providing appropriate system prompts and example passages of classic scenes featuring each character. We generate a dataset of 54K simulated dialogues and demonstrate the feasibility of fine-tuning multiple roles into a single ~7B parameter local model. 

As our first attempted character is the vividly characterized Haruhi Suzumiya, we name the project ChatHaruhi and the dataset Haruhi-54K. Accompanying this report, we release model A trained on 23K original transcripts and model B trained on the full 54K dataset, demos on HuggingFace, and the complete ChatHaruhi-54K dataset.

In future iterations, we will refine the ChatHaruhi interface for easier reusability (see Appendix \ref{sec:AppendixA} for ChatHaruhi 2.0 draft), and supplement the quantitative evaluations.

\section{ Acknowledgments }

This is an open source project originating from the June 2023 study group of the DataWhale community, where we tested an early Haruhi-only version of the chatbot and received highly enthusiastic feedback. We then recruited volunteers from the community (e.g. Yan Chenxi, Feng Xiaoyang) for collaborative development. DataWhale nominated the project for the ModelScope Hackathon in early July, bringing greater exposure and drawing more developers. We sincerely thank DataWhale and ModelScope for their support during this project.

ChatHaruhi is also a sub-project of the open source community \href{https://github.com/LC1332/Luotuo-Chinese-LLM}{Luotuo}, which has benefited from various donations including funding, computing resources, and OpenAI API credits. We thank Luotuo's sponsors for their supports.

\subsection{Contributors}\label{sec:contributor}

This project is an open source work, with all personnel contributing and developing in their spare time. The developers of this project may belong to other institutions or be independent developers. Here we list the main contributions of each developer, as well as their affiliations.

\noindent
\href{https://github.com/LC1332}{Cheng Li}@SenseTime purposed the entire project and designed and implemented most of the functionality.

\noindent
\href{https://blairleng.github.io}{Ziang Leng}@SenseTime designed and implemented the overall training, data generation and backend architecture of ChatHaruhi1.0.

\noindent
\href{https://github.com/todochenxi}{Chenxi Yan}@Chengdu University of Information Technology implemented and maintained the backend of ChatHaruhi1.0 version.

\noindent
\href{https://github.com/J1shen}{Junyi Shen}@Zhejiang University implemented the training code and participated in the generation of training dataset.

\noindent
\href{https://github.com/wanghao07456}{Hao Wang} collected script data from My Own Swordsman and participated in the generation of augmented data.

\noindent
\href{https://github.com/hhhwmws0117}{Weishi MI}@Tsinghua University participated in the generation of augmented data.

\noindent
\href{https://ariafyy.github.io/}{Yaying Fei}@Beijing University of Technology implemented the ASR function of the script tool and participated in the Openness-Aware Personality paper sub-project.

\noindent
\href{https://github.com/fengyunzaidushi}{Xiaoyang Feng}@Nanjing Agricultural University integrated the functions of the script recognition tool and participated in the Openness-Aware Personality paper sub-project.

\noindent
\href{https://github.com/zealot52099}{Song Yan} collected data from The Big Bang Theory. Implemented script format conversion functionality.

\noindent
\href{https://github.com/ssccinng}{HaoSheng Wang} implemented voiceprint recognition function in script tool, and tts-vits speech synthesis function.

\noindent
\href{https://github.com/JunityZhan}{Linkang Zhan}@Case Western Reserve University collected system prompt and story data from Genshin Impact.

\noindent
\href{https://github.com/KaiJiaBrother}{Yaokai Jia} implemented the Vue version of the front end, and practiced GPU extraction of Bert in the psychology project.

\noindent
\href{https://github.com/wpydcr}{Pingyu Wu}@Juncai Shuyun helped deploy the first version of the training code.

\noindent
Haozhen Sun@Tianjin University drew the mosaic of ChatHaruhi characters.

If you have any suggestions for the project, such as the interface design of ChatHaruhi2.0, or want to add references to the future version of this report, please go to our project \href{https://github.com/LC1332/Chat-Haruhi-Suzumiya}{https://github.com/LC1332/Chat-Haruhi-Suzumiya} to submit an issue.

\bibliography{anthology,custom}
\bibliographystyle{acl_natbib}

\appendix

\section{ Appendix-A: ChatHaruhi 2.0 Design } \label{sec:AppendixA}

ChatHaruhi is an open-source conversational agent project. Initially developed for many hackathons, it incorporated multimodal features like images and speech. The project can now be run via a Gradio demo in the source code. However, this design hampers research on multiple chatbots, e.g. adding characters, studying interactions, and upgrading the memory. We will refactor ChatHaruhi after this arXiv paper published, with a tentative API as follows: 

\begin{lstlisting}[style=mystyle]
from ChatHaruhi import ChatHaruhi

chatbot = ChatHaruhi(system_prompt='prompt.txt',

story_db='./story_chroma_folder',
llm='openai')

response = chatbot.chat(text='Can you introduce yourself?',
role='Kyon')
\end{lstlisting}

This simplifies access via a system prompt and vector database. LLM switching also becomes possible, including local models, Claude, and Xinghuo etc. 

If using the characters involved in ChatHaruhi-54K, you can directly use:

\begin{lstlisting}[style=mystyle]
from ChatHaruhi import ChatHaruhi

chatbot = ChatHaruhi(role_name = 'Haruhi',

llm = 'openai')

response = chatbot.chat(text = 'Can you introduce youself?',
role = 'Kyon')
\end{lstlisting}

\end{document}